\pdfoutput=1

\documentclass[11pt,a4paper]{article}
\usepackage[hyperref]{eacl2021}
\usepackage{times}
\usepackage{latexsym}

\usepackage[subrefformat=parens]{subcaption}

\usepackage{microtype}

\aclfinalcopy %
\def\aclpaperid{356} %

\usepackage{color}
\usepackage{colortbl}
\usepackage{tabularx}
\usepackage{amsmath}
\usepackage{amsfonts}
\usepackage{amssymb}
\usepackage{multirow}
\usepackage{here}
\usepackage{setspace}
\usepackage{tikz,tikz-qtree}
\usetikzlibrary{shapes.geometric,positioning,shapes,fit,arrows,backgrounds}
\usetikzlibrary{intersections, calc}
\usepackage{algorithm}
\usepackage{algorithmic}
\usepackage{proof}
\usepackage{ebproof}
\usepackage{textcomp}
\usepackage[T1]{fontenc}
\usepackage{arydshln}

\usepackage{url}
\usepackage{mediabb}
\usepackage{gb4ea}

\newcommand{\modelcite}[1]{\citeauthor{#1},\ \citeyear{#1}}
\newcommand{\fpp}{$f(s_1)\!\rightarrow\!s_1$}
\newcommand{\fph}{$f(s_1)\!\rightarrow\!s_2$}
\newcommand{\f}{$f$}
\newcommand{\fs}{$f(s_1)$}
\newcommand{\fx}{$f(x)$}
\newcommand{\ph}{$s_1\!\rightarrow\!s_2$}
\newcommand{\basicone}{$\mathcal{I}_1$}
\newcommand{\basictwo}{$\mathcal{I}_2$}

\newcommand{\yes}{\textit{yes}}
\newcommand{\unk}{\textit{unk}}
\newcommand{\F}{\texttt{V}}
\newcommand{\NF}{\texttt{NV}}

\newcommand{\diff}[1]{\textcolor{blue}{#1}}

\newcommand{\sem}[1]{[\![ #1 ]\!]}

\newcommand{\catFont}[1]{\textsc{#1}}
\newcommand{\stringFont}[1]{\texttt{#1}}

\newcommand{\strcat}[1]{\ensuremath{\stringFont{#1}}}
\newcommand{\catS}{\ensuremath{\catFont{S}}}
\newcommand{\catSNEG}{\ensuremath{\catFont{Sneg}}}
\newcommand{\catNP}{\ensuremath{\catFont{NP}}}
\newcommand{\catPN}{\ensuremath{\catFont{PN}}}
\newcommand{\catVNEG}{\ensuremath{\catFont{Vneg}}}

\newcommand{\catVPtense}{\ensuremath{\catFont{VP}_{\mathit{tense}}}}
\newcommand{\catVPpast}{\ensuremath{\catFont{VP}_{\text{past}}}}
\newcommand{\catVPbase}{\ensuremath{\catFont{VP}_{\text{base}}}}

\newcommand{\catIV}{\ensuremath{\catFont{IV}}}
\newcommand{\catIVtense}{\ensuremath{\catFont{IV}_{\mathit{tense}}}}
\newcommand{\catIVpast}{\ensuremath{\catFont{IV}_{\text{past}}}}
\newcommand{\catIVbase}{\ensuremath{\catFont{IV}_{\text{base}}}}
\newcommand{\catTVtense}{\ensuremath{\catFont{TV}_{\mathit{tense}}}}
\newcommand{\catTVbase}{\ensuremath{\catFont{TV}_{\text{base}}}}
\newcommand{\catTVpast}{\ensuremath{\catFont{TV}_{\text{past}}}}
\newcommand{\catTV}{\ensuremath{\catFont{TV}}}

\newcommand{\catConn}{\ensuremath{\catFont{CON}}}

\newcommand{\symb}{\ensuremath{\mathbf{sym}}}

\newcommand{\LF}[1]{\ensuremath{\mathbf{#1}}}

\definecolor{LightCyan}{rgb}{0.88,1,1}

\usepackage{tcolorbox}
\newtcbox{\orangebox}{on line,
  colframe=orange,colback=orange!5!white,
  boxrule=0.5mm,
  arc=1.0mm,
  boxsep=0pt,left=1pt,right=1pt,top=1pt,bottom=1pt
  }

\newtcbox{\bluebox}{on line,
  colframe=blue!70!black,colback=blue!5!white,
  boxrule=0.5mm,
  arc=1.0mm,
  boxsep=0pt,left=1.0pt,right=1.0pt,top=1.0pt,bottom=1.0pt
  }
 
\newtcbox{\orangeboxtitle}{on line,
  colframe=white,colback=orange!15!white,
  boxrule=0mm,left=1pt,right=1pt,top=1pt,bottom=1pt
  }

\newtcbox{\blueboxtitle}{on line,
  colframe=white,colback=blue!15!white,
  boxrule=0mm,left=1pt,right=1pt,top=1pt,bottom=1pt
  }
 
\newtcbox{\graybox}{on line, sharp corners,
  colframe=black,colback=gray!10!white,
  boxrule=0.3mm,left=1pt,right=1pt,top=1pt,bottom=1pt
  }

\title{Exploring Transitivity in Neural NLI Models through Veridicality}

\author{
  \parbox{\linewidth}{\centering
   Hitomi Yanaka$^1$,
   Koji Mineshima$^2$, and
   Kentaro Inui$^{3,1}$
  }
  \\
   $^1$\mbox{\rm RIKEN,}
   $^2$\mbox{\rm Keio University,}
   $^3$\mbox{\rm Tohoku University} 
  \\
  \parbox{\linewidth}{\centering
   {\tt hitomi.yanaka@riken.jp},
   {\tt minesima@abelard.flet.keio.ac.jp},
   {\tt inui@ecei.tohoku.ac.jp}
   }
}

\date{}

\begin{document}
\maketitle
\begin{abstract}
Despite the recent success of deep neural networks in natural language processing, the extent to which they can demonstrate human-like generalization capacities for natural language understanding remains unclear. We explore this issue in the domain of natural language inference (NLI), focusing on the \textit{transitivity} of inference relations, a fundamental property for systematically drawing inferences. A model capturing transitivity can compose basic inference patterns and draw new inferences. We introduce an analysis method using synthetic and naturalistic NLI datasets involving clause-embedding verbs to evaluate whether models can perform transitivity inferences composed of veridical inferences and arbitrary inference types. We find that current NLI models do not perform consistently well on transitivity inference tasks, suggesting that they lack the generalization capacity for drawing composite inferences from provided training examples. The data and code for our analysis are publicly available at \url{https://github.com/verypluming/transitivity}.
\end{abstract}

\section{Introduction}
\label{sec:intro}
Deep neural networks (DNNs) have shown impressive performance in many natural language processing tasks.
In particular, DNN models pretrained with large-scale data such as BERT~\cite{BERT2018new} have achieved high accuracy in various benchmark tasks~\cite{NIPS2019_8589,wang2018glue}, which suggests that they might possess some
generalization capacities that are a hallmark of
human cognition.
However, recent analyses~\cite{talmor-berant-2019-multiqa,liu-etal-2019-inoculation,mccoy2019} have shown that high accuracy on a test set drawn from the
same distribution as the training set does not always indicate
that the model has obtained the intended ability, so
it remains unclear
to what extent DNN models
can learn the systematic
generalization
in natural language from
training instances.

\begin{figure}[tb]
\centering
\scalebox{0.75}{
\begin{tikzpicture}

\tikzset{cross/.style={cross out, draw=black, thick, minimum size=2*(#1-\pgflinewidth), inner sep=0pt, outer sep=0pt}}

\node (veridical) at (3.0, 5.7) {\textbf{Veridical inference}};
\draw[blue, thick, rounded corners=10pt] (-0.5,1.5)--(-0.5,6)--(6.5,6)--(6.5,1.5)-- cycle;

\node (boolean) at (5.3, 4.2) {\textbf{Boolean inference}};
\draw[orange, thick, rounded corners=10pt] (1,2.9)--(1,4.5)--(9.2,4.5)--(9.2,2.9)-- cycle;

\node (A) at (3,5) {$A$: Jo knows that Ann and Bob left.};
\node[below=1.0cm of A] (B) {$B$: Ann and Bob left.};
\node[below=1.0cm of B] (Adash) {$A'$: Jo hopes that Ann and Bob left.};
\node[right=2.2cm of B] (C) {$C$: Ann left.};

\draw[->, thick] (A) edge (B);
\draw[->, thick] (Adash) edge (B);
\node[cross=6pt] at (3, 2.6){};
\draw[->, thick] (B) edge (C);

\draw (A.east) edge [->, ultra thick, densely dashed, bend left] (C.north);
\draw (Adash.east) edge [->, ultra thick, densely dashed, bend right] (C.south);
\node[cross=6pt, rotate=40] at (7.45, 2.42){};

\end{tikzpicture}
}

\caption{Illustration of transitivity inferences
(indicated by \raisebox{0.7mm}{\tikz{\draw[very thick,densely dashed,->](0,0) -- (0.4,0)}})
composed of two basic inferences, veridical and Boolean.
Arrows indicate \textit{entailment}
and arrows with a cross (\tikz{\node[cross out, draw=black, thick, minimum size=5, inner sep=0pt, outer sep=0pt]{}})
indicate \textit{non-entailment}.}
\label{fig:goodpic}
\end{figure}

Central to human-like generalization capacities is the fact that ability to understand a given sentence is related to ability to understand other sentences,
called \textit{systematicity} of human cognition in \citet{Fodor1988-FODCAC}.
Thus, if speakers understand the meaning of the sentence \textit{Ann loves Bob}, they must also understand the meaning of structurally related sentences such as \textit{Bob loves Ann}.
We explore whether DNN models possess this type of generalization capacity in the domain of natural language inference (NLI),
which is the task to judge whether a premise entails a hypothesis~\cite{series/synthesis/2013Dagan,snli:emnlp2015}.

A key property underlying systematicity of drawing inferences is the \textit{transitivity} of inference relations, illustrated in Figure \ref{fig:goodpic}.
Schematically, if a model learns a basic inference pattern from $A$ to $B$ and one from $B$ to $C$, it should be able to compose the two patterns to draw a new inference from $A$ to $C$.
If a model lacks this generalization capacity, it must memorize an exponential number of inference combinations
independently of basic patterns.

Among the various inference patterns,
we focus on transitivity inferences that combine \textit{veridical} inferences with
other types.
In veridical inferences, one must distinguish two entailment types.
For example, the verb \textit{know}
is called \textbf{veridical} in that
``$x$ \textit{knows that} $P$'' entails that $P$ is true, while the verb \textit{hope} is called \textbf{non-veridical} since ``$x$ \textit{hopes that} $P$'' does not entail that $P$ is true.
Veridical inferences can 
relatively easily 
compose transitivity inferences at scale by embedding various inference types into clause-embedding verbs.
For instance, as Figure~\ref{fig:goodpic} shows, if a model has the ability to perform both Boolean inference and veridical inference,
it is desirable to have the ability to combine both types to make a chained inference.

Such transitivity inferences are by no means trivial. For instance, 
if the premise is changed to 
\textit{Jo knows that Ann \textbf{or} Bob left}, it does not follow that \textit{Bob left}, even though the veridical verb \textit{know} appears.
Models relying on shallow heuristics such as lexical overlap can wrongly predict \textit{entailment} in this case.
To correctly handle such composite inferences,
models must capture structural relations between veridical inferences
and various kinds of embedded inference.

Previous studies on the generalization capacities of NLI models have addressed how models
could learn inferences with various challenging linguistic phenomena~\cite{Bowman2015,dasgupta2018evaluating,Geiger2019,geiger-etal-2020-neural,yanaka-etal-2019-neural,yanaka2019,Richardson2019}.
However, these studies have focused on the linguistic phenomena in isolation,
and thus do not address how a model could learn the \textit{interactions} between them.
Our aim is to fill this gap by presenting a method for probing generalization capacity of DNN models performing transitivity inferences. 

This study provides three main contributions.
First, we
create and publicly release
two types of NLI datasets for testing model ability to perform transitivity inferences:
a fully synthetic dataset that combines veridical inferences and Boolean inferences, and a naturalistic dataset
that combines veridical inferences with
lexical and structural inferences.
Second, we use these datasets to systematically expose models to basic inference patterns and test them on a variety of combinations.
This will demonstrate that the models lack the ability to capture transitivity of inference.
Third, we investigate whether data augmentation with new combination patterns
helps models to learn transitivity.
Experiments show that the data augmentation improves model performance on similar combinations, regardless of the existence of basic inference patterns in the training set.
These results suggest there is much room for improving the generalization capacities of DNN models for
combining basic inferential abilities.

\section{Related Work}
\label{sec:related}
\paragraph{Transitivity}
The transitivity of entailment relations, which derives $A \to C$ from $A \to B$ and $B \to C$, is incorporated into logic-based NLI systems using automated theorem proving~\cite{abzianidze:2015:EMNLP,D16-1242}.
This is a basic property of formal logic, also known as \textit{syllogism} in traditional logic or the \textit{cut rule} in proof theory~\cite{Troelstra-Schwichtenberg00,vanDalen2013}.
Transitivity inference in its various forms has also been widely studied as a fundamental property of human reasoning in cognitive psychology~\cite{johnson1991deduction,khemlani2012theories}.
In the context of NLP, previous works have proposed a method for training models with transitivity constraints in multi-hop reasoning tasks~\cite{asai-hajishirzi-2020-logic} and temporal relation extraction tasks~\cite{ning-etal-2017-structured}.
\citet{Clark2020} investigated a transformer's ability to perform
a chain of reasoning
where reasoning rules are explicitly given.
In this work, we study model ability to learn transitivity of entailment relations from training examples,
rather than explicitly providing rules.

\paragraph{Systematicity}
There has been extensive discussion of
whether neural networks (aka Connectionist models) can exhibit systematicity of cognitive capacities~\cite{Fodor1988-FODCAC,962dc7dfb35547148019f194381d2cc6}.
Recent works have explored whether modern neural networks can learn systematicity in semantic parsing tasks~\cite{Lake2017GeneralizationWS,baroni2020linguistic,kim-linzen-2020-cogs} and question answering tasks~\cite{sinha-etal-2019-clutrr}, whereas our focus is the systematicity in NLI.

In works related to the systematicity in NLI,
\citet{goodwin-etal-2020-probing},
\citet{yanaka-EtAl:2020:acl},
and \citet{geiger-etal-2020-neural} 
used a manually constructed NLI dataset
of monotonicity inferences with and without negation
(e.g., \textit{The child is not holding plants}
$\to$ \textit{The child is not holding flowers})
to examine DNN models' generalization capacities.
While these approaches concentrate on monotonicity inferences involving quantifiers and negative expressions, our method using veridical inference
is general in that it can be applied to
any entailment relation that combines basic inference patterns;
we generate composite inferences
by embedding various types of sentences into clause-embedding verbs.

\citet{Fodor1988-FODCAC} distinguished
systematicity (roughly, the ability to understand
sentences that are structurally related to each other) from productivity (the ability to understand an infinite set of sentences), claiming
that systematicity poses a serious challenge to neural network models.
\citet{yanaka-EtAl:2020:acl}
tested both systematicity and productivity of DNN models with a synthetic dataset of monotonicity inferences for upward (e.g., \textit{some, at least three})
and downward (e.g., \textit{few, at most three}) quantifiers,
where handling productivity (recursion)
makes sentences more involved (e.g., iterated relative clauses and negation).
Focusing on systematicity rather than productivity allows testing models
with more natural and less complicated data,
as compared to sentences appearing in monotonicity inferences.

\paragraph{Veridicality}
Veridical inferences, including those
licensed by factive and implicative verbs,
have been intensively studied in the literature of semantics and pragmatics~\cite{Karttunen-Peters79,Beaver01}.
Recent work has revealed graded and context-sensitive aspects of veridicality inferences, creating veridicality judgement datasets~\cite{de-marneffe-etal-2012-happen,white2018role,white-etal-2018-lexicosyntactic}.
While we use only a subset of veridical predicates discussed in the literature,
our method can be extended to more complex inferences, such as factive presupposition.

\citet{ross-pavlick-2019-well} presented a naturalistic veridicality dataset and compared
the predictions of a BERT-based NLI model and human judgements.
These previous studies on veridicality inferences 
have tended to focus on relations
between whole sentences (e.g., 
\textit{Jo remembered that there was a wild deer jumping a fence})
and its embedded material (e.g., \textit{There was a wild deer jumping a fence}).
By contrast, we consider the interactions of veridicality inferences and other inference types (see Section~\ref{subsec:creation}), including cases where
the embedded material is further paraphrased via
linguistic phenomena (e.g., \textit{Jo remembered that there was a wild deer jumping a fence} $\Rightarrow$ \textit{An animal was jumping}).
We also collect human judgements on our dataset and compare them with model predictions (see Section~\ref{sec:human}).

\paragraph{Probing NLI models}
Many studies of probing NLI models have found that current models often fail on linguistically challenging (adversarial) inferences~\cite{rozen-etal-2019-diversify,DBLP:conf/aaai/NieWB19,yanaka-etal-2019-neural, Richardson2019}, learning undesired biases~\cite{glockner-shwartz-goldberg:2018:Short,poliak-EtAl:2018:S18-2,DBLP:conf/lrec/Tsuchiya18,liu-etal-2019-inoculation}, and heuristics~\cite{mccoy2019}.
Our approach also provides adversarial test sets against such heuristics by considering combinations of veridical inferences and diverse (lexical, structural, and logical) types of inferences.

One way to learn challenging inferences is data augmentation, and prior studies~\cite{yanaka2019, Richardson2019,min2020syntactic} have shown that data augmentation with synthesized datasets improves performance with challenging linguistic phenomena.
However, it remains unclear whether data augmentation can help models learn \textit{composite} inferences mixing several inference types from training instances.
We address this question in Section~\ref{ssec:dataaug}.

\section{Dataset}
\label{sec:dataset}

\subsection{Overview}
\label{subsec:method}
To investigate whether models can capture transitivity, we consider two basic inference patterns and their combinations.
The first basic pattern, \basicone, is veridical inference.
We write \fpp\ to denote a schematic 
veridical inference, where $f$ is a clause-embedding verb and $s_1$ is the embedded clause.
For instance, in the case of the inference pattern
$A\rightarrow B$
in Figure~\ref{fig:goodpic}, ``\textit{Jo knows that x}'' corresponds to \fx\ and ``\textit{Ann and Bob left}'' to $s_1$.

\begin{table*}[tb]
\centering
\scalebox{0.80}{
\begin{tabular}[t]{lllll} \hline 
\f&\fpp&\ph&\fph&\textbf{Example}\\ \hline
\F&\yes&\yes&\yes&\begin{tabular}{p{32em}} \fs: Someone \textbf{noticed} that [Henry and Daniel found Elliot, John and Fred].\\
$s_1$: Henry and Daniel found Elliot, John and Fred.\\
$s_2$: Henry found John.
\end{tabular}\\ \hline
\NF&\unk&\yes&\unk&\begin{tabular}{p{32em}} \fs: Someone \textbf{expects} that [Tom and Ann admire Greg and Fred].\\
$s_1$: Tom and Ann admire Greg and Fred.\\
$s_2$: Tom admires Greg.
\end{tabular}\\ \hline
\NF&\unk&\unk&\unk&\begin{tabular}{p{32em}} \fs: Someone \textbf{argued} that [it was not the case that Greg hated John or Elliot].\\
$s_1$: It was not the case that Greg hated John or Elliot.\\
$s_2$: Greg hated John.
\end{tabular}\\ \hline
 \end{tabular}
}
  \caption{Examples from our fully synthetic transitivity inference datasets. \F\ and \NF\ indicate types of clause-embedding verbs (veridical/non-veridical); \yes\ means \textit{entailment} and \unk\ means \textit{non-entailment}.}
 \label{tab:full-examples}
\end{table*}

\begin{table*}[tb]
\centering
\scalebox{0.80}{
\begin{tabular}[t]{llllll} \hline 
\textbf{ID}&\f&\fpp&\ph&\fph&\textbf{Example}\\ \hline
2299&\F&\yes&\yes&\yes&\begin{tabular}{p{29em}} \fs: Someone \textbf{realized} that [a boy was playing a guitar].\\
$s_1$: A boy was playing a guitar.\\
$s_2$: A kid was playing a guitar.
\end{tabular}\\ \hline
2049&\F&\yes&\unk&\unk&\begin{tabular}{p{29em}} \fs: Someone \textbf{remembered} that [a cat was playing with a device].\\
$s_1$: A cat was playing with a device.\\
$s_2$: The boy was enthusiastically playing in the mud.\
\end{tabular}\\ \hline
5024&\NF&\unk&\yes&\unk&\begin{tabular}{p{29em}} \fs: Someone \textbf{doubts} that [the woman is putting makeup on the man].\\
$s_1$: The woman is putting makeup on the man.\\
$s_2$: A man's face is being painted by a woman. 
\end{tabular}\\ \hline
 \end{tabular}
}
  \caption{Examples from our naturalistic transitivity inference datasets. \F\ and \NF\ indicate types of clause-embedding verbs (veridical/non-veridical); \yes\ means \textit{entailment} and \unk\ means \textit{non-entailment}. \textbf{ID} indicates the original ID of \ph\ in the SICK dataset.}
 \label{tab:semi-examples}
\end{table*}

The second basic pattern, \basictwo,
provides an inference from the embedded material.
We denote a premise-hypothesis pair of this second inference by \ph.
Given two inferences \fpp\ in \basicone\ and \ph\ in \basictwo,
we consider a new inference \fph,
where premise \fs\ is the same as that of $\mathcal{I}_1$ and hypothesis $s_2$ is the same as that of $\mathcal{I}_2$.
See Table~\ref{tab:full-examples} and Table~\ref{tab:semi-examples} for some examples of inferences \fpp, \ph, and \fph.
In this work, we consider binary labels,
\textit{entailment} and \textit{non-entailment},
denoted by \yes\ and \unk, respectively.
As Table~\ref{tab:labels} shows, the gold label on the \fph\ pattern can be determined from those of the basic patterns \fpp\ and \ph,
following the transitivity of entailment relations.

\begin{table}[!ht]
\centering
\scalebox{0.95}{
\begin{tabular}[t]{cc|c} \hline 
\fpp&\ph&\fph \\ \hline
\yes&\yes&\textbf{\yes} \\
\yes&\unk&\textbf{\unk} \\
\unk&\yes&\textbf{\unk} \\
\unk&\unk&\textbf{\unk} \\ \hline
\end{tabular}
}
\caption{Rule for determining the \fph\ label from the basic patterns \fpp\ and \ph.}
\label{tab:labels}
\end{table}

We train models with
the first and second patterns, \fpp\ and \ph,
and then test them on a set of the composite inferences \fph\ that combines them.
Note that due to how they are constructed, the training and test sets do not overlap.
Model capable of applying the transitivity inference from \fpp\ and \ph\ to \fph\ should consistently predict the correct label of \fph\ for any combination of \fpp\ and \ph.

\subsection{Data creation}
\label{subsec:creation}
We generate basic inferences
\fpp\ and \ph\
and combine them to produce
transitivity inferences
\fph.
To test diverse inference patterns,
we consider two types of the second basic inference \ph:
synthesized Boolean inferences and
naturalistic inferences using an existing NLI dataset, SICK~\cite{marelli-etal-2014-sick},
which contains lexical inferences (e.g., \textit{boy} $\to$ \textit{kid} in ID 2299 in Table~\ref{tab:semi-examples}) and structural inferences (e.g., active-passive alternation in ID 5024 in Table~\ref{tab:semi-examples}).
Since the ratio of the gold labels (\yes\ and \unk) is set to $1:1$ in both
basic inference sets, the ratio of the gold labels for the transitivity test set is $1:3$
by the rule in Table~\ref{tab:labels}.
We reserve 10\% of the basic inference set for the validation set.

\begin{table}[tb]
    \centering
    \scalebox{0.87}{
    \begin{tabular}{lp{15em}}\\ \hline
    \textbf{Type of \f}&\textbf{Verbs}\\ \hline
    Veridical&realize, acknowledge, remember, note, find, notice, learn, see, reveal, discover, understand, know, admit, recognize, observe\\
    Non-veridical&feel, claim, doubt, hope, predict, imply, suspect, wish, think, believe, hear, expect, estimate, assume, argue \\ \hline
    \end{tabular}
    }
    \caption{Clause-embedding verbs used for our dataset.}
    \label{tab:verbs}
\end{table}

\paragraph{Clause-embedding verbs}
We focus on clause-embedding verbs that take tensed subordinate clauses.
Specifically, we collect 67 verbs appearing in both MegaVeridicality2~\cite{white-etal-2018-lexicosyntactic} and the verb veridicality dataset~\cite{ross-pavlick-2019-well}.
As Table~\ref{tab:verbs} shows, we select a final set of 30 clause-embedding verbs.

Following a previous study~\cite{white-etal-2018-lexicosyntactic}, we slot a clause-embedding verb \f\ into a template with the form ``Someone \f\ that $s_1$'' and
generate premise \fs\ of veridical inference
to avoid confounds introduced by world knowledge and pragmatic inference in the main clause.
The clause-embedding verb \f\ is in past or present tense, and we inflect the verb in the complement $s_1$ to match the tense of \f.

When measuring the extent to which models can learn transitivity of entailment relations from training instances, it is desirable to determine the gold labels of composite inferences from those of basic inferences.
Thus, we take the labels of veridical inference datasets predicted by the veridical and non-veridical distinction in lexical semantics as the gold standard.
In addition, veridical inferences are sensitive to context, influenced by world knowledge and pragmatic factors~\cite{de-marneffe-etal-2012-happen}.
Accordingly, we also present additional experiments to take into account such complexity of veridical inferences in Section~\ref{ssec:trans}.

\paragraph{Boolean inference}
To provide a fully synthetic transitivity inference dataset, we generate Boolean inferences with conjunction, disjunction, and negation.
The data generation process is similar to the one in
\citet{yanaka-EtAl:2020:acl}: sentences are generated using a context-free grammar (CFG) associated with semantic composition rules in lambda-calculus.
We first generate a set of premise sentences by the CFG rules and translate each sentence $s_1$ into a first-order-logic (FOL) formula $F_1$ in accordance with semantic composition rules specified in the CFG rules.
Appendix A provides a set of CFG rules and semantic composition rules.
We randomly select one of the atomic sub-formulas appearing in $F_1$ and take its positive or negative form, which we denote by $F_2$.
Then we convert $F_2$ to a sentence $s_2$ using the same grammar.
We set $s_2$ as a hypothesis.

The gold label for inference pair \ph\ is determined by checking whether formula $F_1$ entails formula $F_2$ using an FOL theorem prover.
The gold labels for \fpp\ and \fph\ pairs are automatically determined according to the veridicality of a clause-embedding verb and the rule in Table~\ref{tab:labels}, respectively.
To restrict the complexity of generated sentences,
we set the maximum number of logical connectives appearing in formula $F_1$ to 6.

Table~\ref{tab:full-examples} illustrates examples of fully synthetic transitivity inference datasets.
We generate 3,000 Boolean inference examples \ph,\ 6,000 veridical inference examples \fpp,\ and 6,000 composite inference examples \fph.

\paragraph{Naturalistic inference}
To generate a naturalistic transitivity inference dataset, we collect an example \ph\ of naturalistic inference from the SICK dataset, which is
constructed from existing sentences 
(image descriptions
given by different people) and covers various lexical and structural phenomena.
(\ref{ex:1}) is an example of lexical inference (\textit{brush} $\to$ \textit{comb}) in SICK,
whose label is \yes.

\begin{exe}
\ex \label{ex:1}
\begin{xlist}   
\exi{$s_1$:} \label{ex:1a} 
A person is brushing a cat.
\exi{$s_2$:}\label{ex:1b} 
A person is combing the fur of a cat. 
\end{xlist}
\end{exe}

\noindent
By selecting a clause-embedding verb \f\ and an embedded sentence $s_1$,
we generate a new sentence \fs.
As shown in (\ref{ex:2}), we construct a veridical inference example \fpp\ by setting \fs\ as a premise and $s_1$ as a hypothesis.
\begin{exe}
\ex \label{ex:2}
\begin{xlist}
\exi{\fs:} \label{ex:2a}
Someone \textbf{sees} that a person is brushing a cat.
\exi{$s_1$:} \label{ex:2b}
A person is brushing a cat. 
\hfill{(\yes)}
\end{xlist}
\end{exe}
Likewise, as in (\ref{ex:3}), we can obtain a composite inference example \fph\ whose label is \yes:
\begin{exe}
\ex \label{ex:3}
\begin{xlist}
\exi{\fs:} \label{ex:3a}
Someone \textbf{sees} that a person is brushing a cat.
\exi{$s_2$:} \label{ex:3b}
A person is combing the fur of a cat. 
\end{xlist}
\end{exe}

Table~\ref{tab:semi-examples} illustrates examples of  naturalistic transitivity inference datasets.
We sample 1,000 naturalistic inference examples \ph\  from the SICK training set and obtain 30,000 veridical inference examples \fpp\ and 30,000 composite inference examples \fph.

\begin{table*}[tb]
\centering
\scalebox{0.81}{
\begin{tabular}[t]{cc|c|cccccc} \hline
\multicolumn{3}{c|}{\textbf{Data}} &\multicolumn{6}{c}{\textbf{Model}} \\
\fpp&\ph&\fph& \textbf{LSTM-M} & \textbf{LSTM-B} & \textbf{LSTM-M\&B} & \textbf{BERT-M} & \textbf{BERT-B} & \textbf{BERT-M\&B}\\ \hdashline
\yes&\yes&\textbf{\yes}&$74.2\pm2.0$&$89.0\pm9.1$&$87.9\pm3.7$&$66.3\pm3.4$&$100.0\pm0.0$&$100.0\pm0.0$ \\
\rowcolor{LightCyan}
\yes&\unk&\textbf{\unk}&$16.0\pm4.3$&$6.3\pm12.8$&$60.0\pm10.2$&$4.9\pm1.5$&$0.4\pm0.7$&$60.5\pm0.6$ \\ 
\unk&\yes&\textbf{\unk}&$14.7\pm3.8$&$93.4\pm8.3$&$89.0\pm9.5$&$12.6\pm4.8$&$99.4\pm9.0$&$92.9\pm3.6$ \\
\unk&\unk&\textbf{\unk}&$17.8\pm5.5$&$92.1\pm7.2$&$99.7\pm0.5$&$13.2\pm3.4$&$99.5\pm0.5$&$99.9\pm0.0$ \\ 
\hdashline
\multicolumn{3}{c|}{\textbf{Test Overall}}&$30.9\pm3.2$&$70.2\pm3.4$&$84.2\pm1.2$&$24.4\pm1.6$&$75.7\pm0.4$&$88.3\pm0.9$\\
\hline
\multicolumn{3}{c|}{\textbf{Validation} (\fpp)}&$50.5\pm1.7$&$93.3\pm11.1$&$91.4\pm5.7$&$68.1\pm1.3$&$99.2\pm0.2$&$98.3\pm0.3$\\
\multicolumn{3}{c|}{\textbf{Validation} (\ph) }&$41.5\pm3.4$&$89.2\pm3.4$&$85.2\pm1.2$&$54.4\pm2.3$&$100.0\pm0.0$&$99.4\pm0.5$\\
\hline
 \end{tabular}
}
  \caption{Accuracies for the fully synthetic transitivity test set and the validation set. \textbf{-B} indicates a model trained with the basic inference set, \textbf{-M} indicates a model trained with MNLI, and \textbf{-M\&B} indicates a model trained with MNLI mixed with the basic inference set. The label \yes\ means \textit{entailment}, and \unk\ means \textit{non-entailment}.}
 \label{tab:propexp_1}
\end{table*}

\begin{table*}[tb]
\centering
\scalebox{0.81}{
\begin{tabular}[t]{cc|c|cccccc} \hline
\multicolumn{3}{c|}{\textbf{Data}}&\multicolumn{6}{c}{\textbf{Model}} \\
\fpp&\ph&\fph& \textbf{LSTM-M} & \textbf{LSTM-B} & \textbf{LSTM-M\&B} & \textbf{BERT-M} & \textbf{BERT-B} & \textbf{BERT-M\&B}\\ 
\hdashline
\yes&\yes&\textbf{\yes}&$64.6\pm12.1$&$97.1\pm2.7$&$100.0\pm0.1$&$85.9\pm1.1$&$100.0\pm0.0$&$100.0\pm0.0$ \\
\rowcolor{LightCyan}
\yes&\unk&\textbf{\unk}&$45.6\pm10.5$&$0.0\pm0.0$&$3.6\pm1.4$&$28.4\pm0.9$&$8.9\pm7.8$&$22.3\pm13.6$ \\ 
\unk&\yes&\textbf{\unk}&$24.4\pm12.1$&$97.1\pm2.7$&$99.7\pm0.5$&$13.3\pm1.7$&$100.0\pm0.0$&$100.0\pm0.0$ \\
\unk&\unk&\textbf{\unk}&$45.4\pm11.2$&$97.3\pm2.6$&$99.9\pm0.1$&$31.1\pm0.9$&$100.0\pm0.0$&$100.0\pm0.0$ \\
\hdashline
\multicolumn{3}{c|}{\textbf{Test Overall}}&$45.0\pm5.5$&$72.9\pm2.0$&$75.8\pm0.5$&$39.7\pm0.2$&$77.2\pm2.0$&$80.6\pm3.4$\\
\hline
\multicolumn{3}{c|}{\textbf{Validation} (\fpp)}&$46.2\pm1.2$&$82.1\pm3.3$&$89.8\pm6.5$&$68.7\pm1.6$&$99.2\pm0.0$&$97.1\pm0.3$\\
\multicolumn{3}{c|}{\textbf{Validation} (\ph)}&$58.0\pm1.0$&$81.9\pm3.0$&$82.1\pm1.4$&$62.0\pm1.0$&$89.1\pm2.0$&$91.0\pm0.0$\\
\hline
 \end{tabular}
}
\caption{Accuracies for the naturalistic transitivity test set and the validation set.}
\label{tab:lexexp_1}
\end{table*}

\section{Experiments and Analysis}
\label{sec:experiment}
We analyze whether models trained with the basic inference set can consistently perform composite inferences on the test set.
We use two DNN models, BERT and LSTM,
which are known to perform well with
linguistic phenomena such as subject-verb agreement and
hierarchical and structural probing tasks~\cite{linzen-etal-2016-assessing,weiss-etal-2018-practical,kuncoro-etal-2018-lstms}.

\subsection{Experimental setup}
\label{ssec:setup}
In all experiments, 
we train each model for 25 epochs or until convergence and select the best-performing model based on its accuracy on the validation set.
We perform five runs 
and report the average and standard deviation of their accuracies.
\paragraph{LSTM}
We use an LSTM~\cite{Hochreiter:1997:LSM:1246443.1246450} model, where each premise and hypothesis is processed as a sequence of words using RNN with LSTM cells, and the final hidden state of each serves as its representation.
The model concatenates the premise and hypothesis representations and passes the result to three hidden layers followed by a two-way softmax classifier.
The model is initialized with
300-dimensional GloVe vectors~\cite{pennington-etal-2014-glove} and optimized using Adam~\cite{Adam}.
We search dropout probabilities of $[0, 0.1, 0.2]$ on the output.

\paragraph{BERT}
We use the base-uncased pretrained BERT~\cite{BERT2018new} model\footnote{We use the Pytorch implementation of BERT released at https://github.com/huggingface/transformers.}, fine-tuned for the NLI classification task on training data in the standard way. 
When fine-tuning BERT, we search dropout probabilities of $[0, 0.1, 0.2]$ on the output, and hyperparameters are the same as those commonly used for MultiNLI.

\subsection{Testing transitivity}
\label{ssec:trans}
We first evaluate whether the models trained with basic inferences \fpp\ and \ph\ can
consistently make judgements on the composite inferences \fph.
As a previous work~\cite{ross-pavlick-2019-well} reported that a BERT model trained with the benchmark NLI dataset MultiNLI (MNLI;~\modelcite{DBLP:journals/corr/WilliamsNB17})
is sensitive to verb veridicality,
we regard the accuracy of models trained with MNLI as a baseline.
We also analyze
models trained with MNLI mixed with the basic inference set.

Table~\ref{tab:propexp_1} shows accuracies for the fully synthetic transitivity test set
that combines veridical and Boolean inferences.
Models trained with the basic inference set
achieved over 80\% accuracy on the test cases, except for cases where \fpp\ is \yes\ and \ph\ is \unk.
Table~\ref{tab:lexexp_1} shows accuracies for the naturalistic transitivity test set.
Again, models trained with the basic inference set
performed substantially below chance for the
cases \fph,\ where \fpp\ is \yes\ and \ph\ is \unk.
This suggests that while the models achieve over 80\% accuracy on both \fpp\ and \ph\ validation sets,
they do not apply transitivity inference from the inferences \fpp\ and \ph,
but rather predict the label for the composite inference \fph\ by judging whether it is similar to the veridical inference \fpp\ in the training set.

Accuracy of models trained with MNLI was low because they
predicted \yes\ for many examples where correct labels were \unk,\ as in (\ref{ex:res}).
\begin{exe}
\ex \label{ex:res}
\begin{xlist}   
\exi{\fs:} \label{ex:resa} 
Someone wished that \textbf{John saw Tom} or Greg.
\exi{$s_2$:}\label{ex:resb} 
John saw Tom.
\hfill{(\unk)}
\end{xlist}
\end{exe}
The models predicted \yes\ for over 80\% of the fully synthetic transitivity test set and more than 60\% of the naturalistic transitivity test set.
These results are consistent with the findings in \citet{mccoy2019}, namely, that models trained with MNLI tend to predict entailment relations when the hypothesis is a subsequence of the premise, as in (\ref{ex:res}).

When models are trained with MNLI mixed with the basic inference set, they seem to 
improve performance on the fully synthetic transitivity test set.
One reason for this result is that the models might use heuristics to make predictions for some \unk\ examples in the fully synthetic inference set.
Error analysis shows that the models tend to predict \unk\ when either a premise or a hypothesis contains a negation like (\ref{ex:properr}).
\begin{exe}
\ex \label{ex:properr}
\begin{xlist}
\exi{\fs:} \label{ex:properra}
Someone knew that Fred praised Henry or Ann.
\exi{$s_2$:} \label{ex:properrb}
Fred did \textbf{not} praise Ann. 
\hfill{(\unk)}
\end{xlist}
\end{exe}
These heuristics might be related to the annotation artifact~\cite{gururangan-EtAl:2018:N18-2} in MNLI, because an inference example involving negation words tends to be a \textit{contradiction}\footnote{We use binary labels (\textit{entailment}/\textit{non-entailment}) and take \textit{contradiction} as \textit{non-entailment}.}.
Moreover, models can memorize the basic inference set regardless of the existence of MNLI in the training set, so performance seems to be better.

Note that models trained with MNLI mixed with the basic inference set still failed on the naturalistic transitivity inference \fph\ where \fpp\ is \yes\ and \ph\ is \unk.
Since the naturalistic basic inference examples \ph\ contain various linguistic
phenomena, models cannot rely on the heuristics for such examples.

\begin{table}[tb]
    \centering
    \scalebox{0.87}{
    \begin{tabular}{lp{14em}}\\ \hline
    \textbf{Type}&\textbf{Templates}\\ \hline
    Pronoun&At that moment, we \f\ that $s$\\
    Pronoun&Then he \f\ that $s$\\
    Specific group&The customers \f\ that $s$\\
    Specific group&Some economists \f\ that $s$\\
    Proper noun&Hanson \f\ that $s$\\
    \hline
    \end{tabular}
    }
    \caption{Examples of additional templates used for generating veridical inference datasets. Here \f\ is a place for a veridical verb and $s$ for an embedded sentence.}
    \label{tab:templates}
\end{table}

\begin{table}[tb]
\centering
\scalebox{0.68}{
\begin{tabular}[t]{cc|c|cc} \hline
\multicolumn{3}{c|}{\textbf{Data}} &\multicolumn{2}{c}{\textbf{Model}} \\
\fpp&\ph&\fph&\textbf{LSTM-B} ($\triangle{}$)&\textbf{BERT-B} ($\triangle{}$)\\ \hline
\yes&\yes&\textbf{\yes}&$97.9\ (+0.8)$&$100.0\ (0.0)$ \\
\rowcolor{LightCyan}
\yes&\unk&\textbf{\unk}&$0.0\ (0.0)$&$2.3\ (-6.6)$ \\ 
\unk&\yes&\textbf{\unk}&$99.0\ (+1.9)$&$100.0\ (0.0)$ \\
\unk&\unk&\textbf{\unk}&$99.2\ (+1.9)$&$100.0\ (0.0)$ \\
\hline
 \end{tabular}
}
\caption{Accuracies of models in the setting (I).
($\triangle{}$) is the difference from the accuracy in Table~\ref{tab:lexexp_1}.}
\label{tab:various_tmp}

\scalebox{0.68}{
\begin{tabular}[t]{cc|c|cc} \hline
\multicolumn{3}{c|}{\textbf{Data}} &\multicolumn{2}{c}{\textbf{Model}} \\
\fpp&\ph&\fph &\textbf{LSTM-B} ($\triangle{}$)&\textbf{BERT-B} ($\triangle{}$)\\ \hline
\yes&\yes&\textbf{\yes}&$90.0\ (-7.1)$&$93.6\ (-6.4)$ \\
\rowcolor{LightCyan}
\yes&\unk&\textbf{\unk}&$2.2\ (+2.2)$&$17.9\ (+9.0)$ \\ 
\unk&\yes&\textbf{\unk}&$89.9\ (-7.2)$&$94.0\ (-6.0)$ \\
\unk&\unk&\textbf{\unk}&$98.3\ (+1.0)$&$95.6\ (+1.8)$ \\
\hline
 \end{tabular}
}
\caption{Accuracies of models in the setting (II).
($\triangle{}$) is the difference from the accuracy in Table~\ref{tab:lexexp_1}.}
\label{tab:fluctuate}
\end{table}

\begin{figure*}[tb]
\begin{minipage}{0.5\hsize}
\subcaption{\fpp, \ph, and a subset of \fph}
\scalebox{0.39}{\includegraphics[bb=0.000000 0.000000 576.001152 216.000432]{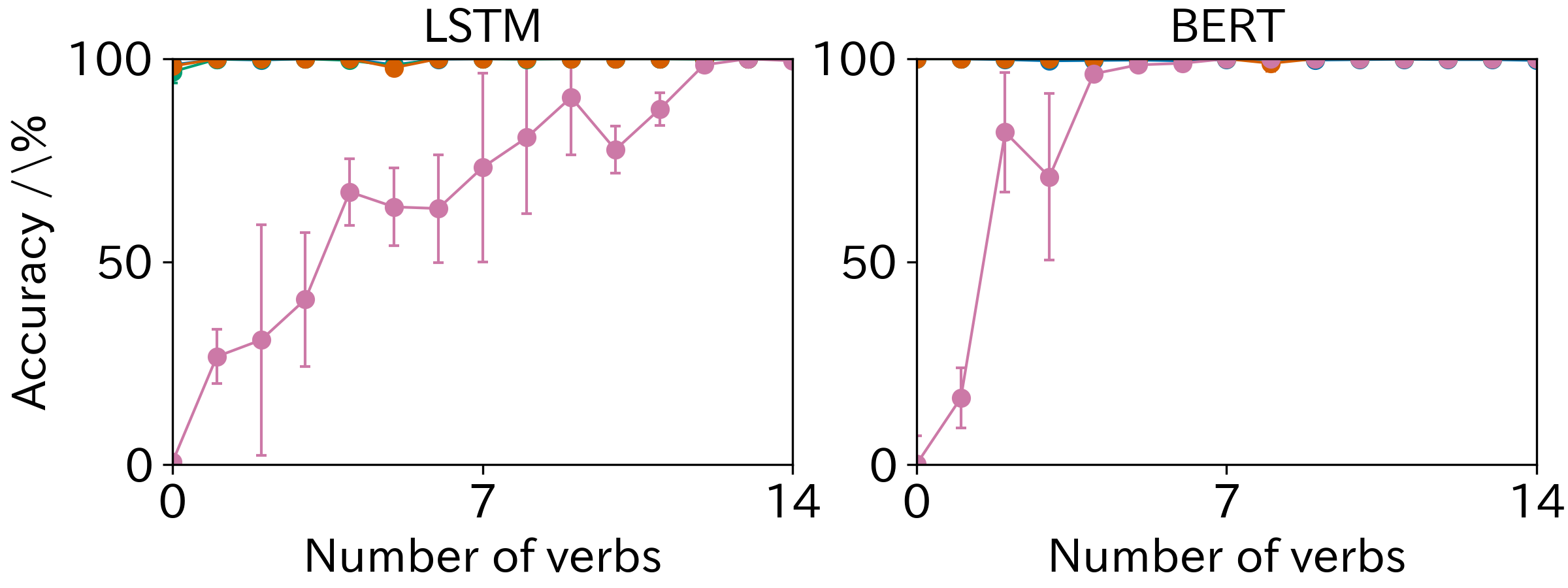}}
\end{minipage}
\begin{minipage}{0.5\hsize}
\subcaption{\fpp\ and a subset of \fph}
\scalebox{0.39}{\includegraphics[bb=0.000000 0.000000 576.001152 216.000432]{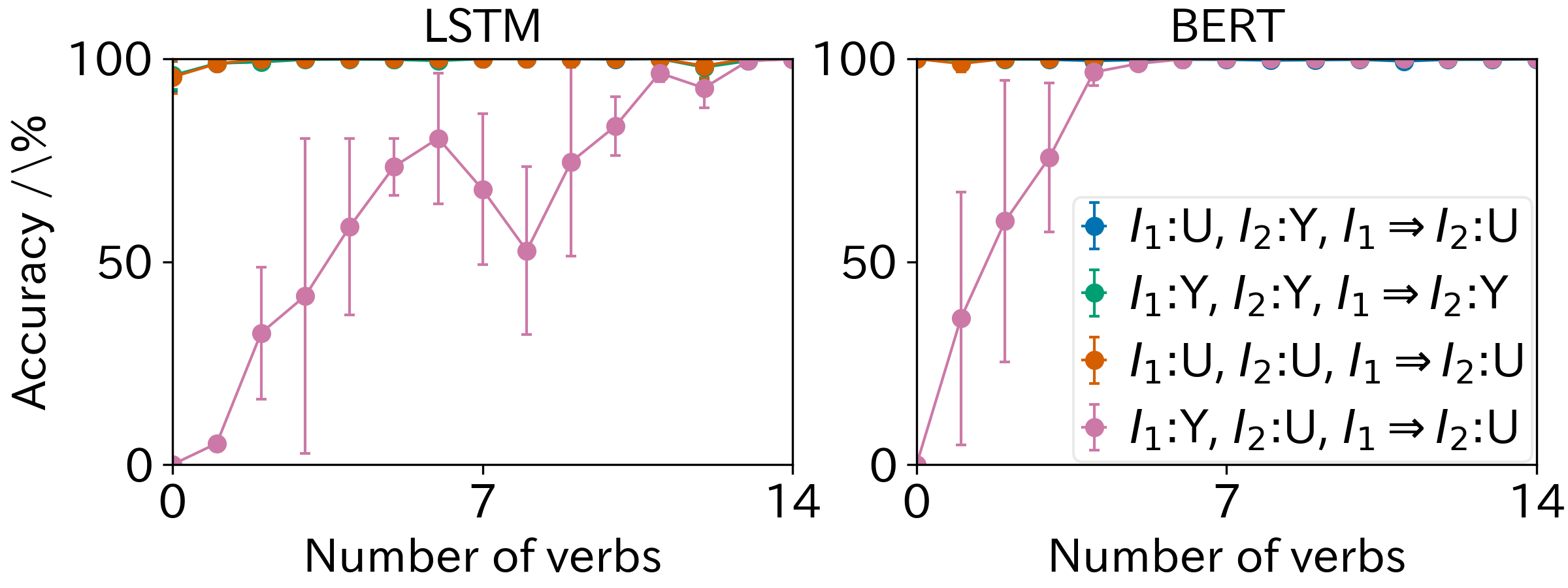}}
\end{minipage}
 \caption{Accuracies of models trained with (a) and (b). $I_1$ indicates the first basic pattern \fpp\ and $I_2$ indicates the second basic pattern \ph. Y means \textit{entailment} and U means \textit{non-entailment}. The horizontal axis shows the number of veridical verbs for the additional training set.}
 \label{fig:part}
\end{figure*}

\begin{table*}[tb]
\centering
\scalebox{0.83}{
\begin{tabular}[t]{cc|c|cc|c} \hline
\multicolumn{3}{c|}{\textbf{Data}} &\multicolumn{2}{c}{\textbf{Model}}&\multirow{2}{*}{\textbf{Human}} \\
\fpp&\ph&\fph & \textbf{LSTM-B} & \textbf{BERT-B}&\\ \hline
\yes&\yes&\textbf{\yes}&$100.0\pm2.7$&$100.0\pm0.0$&$98.8$ \\
\rowcolor{LightCyan}
\yes&\unk&\textbf{\unk}&$0.0\pm0.0$&$8.9\pm7.8$&$98.8$ \\ 
\unk&\yes&\textbf{\unk}&$97.1\pm2.7$&$100.0\pm0.0$&$44.9$ \\
\unk&\unk&\textbf{\unk}&$97.3\pm2.6$&$100.0\pm0.0$&$99.6$ \\
\hdashline
\multicolumn{3}{c|}{\textbf{Test Overall}}&$72.9\pm2.0$&$77.2\pm2.0$&$85.5$\\
\hline
 \end{tabular}
}
  \caption{Comparison between accuracies of humans and the models trained with the basic inference set.}
 \label{tab:human}
\end{table*}

\paragraph{Is poor performance of transitivity inference due to overfitting on verbs?}
To determine whether models do not overfit on clause-embedding verbs, we analyze the models under two additional settings using naturalistic transitivity datasets: (I)~we use various templates
other than ``Someone $f$ that $s_1$''
to generate the main clause in $f(s_1)$, and
(II)~we flip the gold labels of 10\% veridical inference \fpp\ instances, randomly sampled, instead of using gold labels uniquely fixed from verb types.
These two complex settings expose models to more natural evaluation settings that consider the context-sensitive property of veridicality.

For evaluation setting~(I) using various templates involving clause-embedding verbs, we manually select forty main clauses of the verb veridicality dataset~\cite{ross-pavlick-2019-well} and provide additional templates.
Table~\ref{tab:templates} shows examples of additional templates involving clause-embedding verbs used for generating veridical inference datasets.

Table~\ref{tab:various_tmp} and Table~\ref{tab:fluctuate}
show the results for (I) and (II), respectively.
These results show the same trends as those in Table~\ref{tab:lexexp_1},
indicating that even when we consider the complexity of veridical inference in our analysis, the models fail to consistently perform composite inferences.

\subsection{Analysis with data augmentation}
\label{ssec:dataaug}

We further hypothesize that even if the current models fail to consistently perform composite inferences,
data augmentation with a small number of composite inference examples might allow models to learn transitivity inference.
Thus, we evaluate models trained with basic inferences \fpp\ and \ph\ and with a subset of the composite inferences \fph\ 
on a naturalistic inference test set.
Considering that models fail on composite inference \fph\ where \f\ is veridical and \ph\ is \unk,
we gradually add veridical verbs (e.g., \textit{know}) one-by-one to generate an additional training set of composite inference \fph\ and analyze performance on a test set.
Figure~\ref{fig:part}(a) shows that this data augmentation improved performance on test examples \fph\ where \f\ is veridical and \ph\ is \unk,\ while maintaining accuracy on the remaining examples in the test set.
BERT achieved 100\% accuracy over the entire test set by adding composite inferences generated from four veridical verbs, whereas in the case of LSTM twelve veridical verbs were needed to achieve the same accuracy.

To determine whether models augmented with composite inference examples learn the ability to combine basic inferences to perform transitivity inference,
we analyze the performance of models
where basic inference examples are not included in the training set.
Figure~\ref{fig:part}(b) shows that models trained with only the basic inference set \fpp\ and a subset of the composite inference set \fph\ also had improved accuracy.
This result supports findings that models do not combine the basic
inference \fpp\ and \ph,
but rather predict the label for a composite inference \fph\ by judging whether it is similar to inference patterns found in the training set.

\subsection{Comparison with humans}
\label{sec:human}
To investigate how humans perform on transitivity inference tasks, we collect human judgements for a subset of our naturalistic inference dataset.
We asked crowdsourced workers to label 960 transitivity inference examples
involving all the clause-embedding verbs in Table~\ref{tab:verbs}.
Following prior works involving crowdsourced NLI datasets~\cite{zhang-etal-2017-ordinal,ross-pavlick-2019-well}, we instructed raters to label each premise-hypothesis pair with the degree of entailment
on a 5-point Likert scale, with
1 meaning a hypothesis is definitely not true given the premise, and
5 meaning a hypothesis is definitely true.
We collected three annotations per pair on Amazon Mechanical Turk (see Appendix~D for details), and
the inter-rater agreement (the Pearson correlation among raters, averaged across both examples and raters) was 0.76.
As model predictions are discrete 
(\yes\ or \unk), we discretized human scores into evenly sized bins, setting \yes\ if the score was 4 or higher and set
\unk\ if the score was 3 or lower.
We assumed the majority of three discretized labels as the final human judgement.

Table~\ref{tab:human} shows that humans generally follow the distinction between veridical and non-veridical verbs traditionally assumed in the lexical semantics, as well as the transitivity of entailment relation.
In particular, 
while as we saw in Section~\ref{ssec:trans} the DNN models performed substantially below chance
for transitivity inferences where \fpp\ is \yes\ and \ph\ is \unk, human performance is near perfect for such inferences.

Interestingly, however, humans tend to predict incorrect labels for transitivity inferences where the verb \f\ is non-veridical (so \fpp\ is \unk) and the embedded inference \ph\ 
is \yes.
This might be because a natural complement as in (\ref{ex:sickerr}) induces veridicality bias~\cite{ross-pavlick-2019-well}, that is, no matter whether a complement verb $f$ is veridical or non-veridical, humans tend to decide the truth value of \fs\ by judging whether its complement $s_1$ is true.
Thus, judgement for \fph\ coincides with that of \ph\ in this case.

\begin{exe}
\ex \label{ex:sickerr}
\begin{xlist}
\exi{\fs:} \label{ex:sickerr1}
Someone \textbf{believed} that a man is jumping off a low wall.
\exi{$s_1$:} \label{ex:sickerr2}
A man is jumping off a low wall.
\exi{$s_2$:} \label{ex:sickerr3}
A man is jumping a wall.
\end{xlist}
\end{exe}

\section{Conclusion}
\label{sec:conc}
We introduced an analysis method using transitivity inferences for evaluating systematic generalization capacities of NLI models.
We found that current NLI models do not perform consistently well on transitivity inference tasks.
Furthermore, data augmentation analysis suggested
that models can memorize composite inference examples, but do not perform the intended transitivity inferences combining basic inference examples.

Overall, our results indicated that despite the impressive performance of DNN models
on standard NLI datasets, 
there remains much room for improving their systematic generalization capacities with respect to combining basic inferential abilities on various linguistic phenomena.
Regarding what is necessary for improving the systematic generalization capacity, one interesting possibility is explicitly feeding some form of logic-guided transitivity rules to models,
which is left for future work.
Our analysis method using transitivity
can be an effective tool for further progress in the study of compositional NLI.

\section*{Acknowledgement}
We thank the three anonymous reviewers for their helpful comments and suggestions.
We are also grateful to Masashi Yoshikawa for helpful discussions.
This work was partially supported by the RIKEN-AIST Joint Research Fund (feasibility study) and JSPS KAKENHI Grant Number JP20K19868.

\bibliographystyle{acl_natbib}
\bibliography{eacl2021}

\begin{thebibliography}{56}
\expandafter\ifx\csname natexlab\endcsname\relax\def\natexlab#1{#1}\fi

\bibitem[{Abzianidze(2015)}]{abzianidze:2015:EMNLP}
Lasha Abzianidze. 2015.
\newblock A tableau prover for natural logic and language.
\newblock In \emph{Proceedings of the 2015 Conference on Empirical Methods in
  Natural Language Processing (EMNLP)}, pages 2492--2502.

\bibitem[{Asai and Hajishirzi(2020)}]{asai-hajishirzi-2020-logic}
Akari Asai and Hannaneh Hajishirzi. 2020.
\newblock Logic-guided data augmentation and regularization for consistent
  question answering.
\newblock In \emph{Proceedings of the 58th Annual Meeting of the Association
  for Computational Linguistics (ACL)}, pages 5642--5650.

\bibitem[{Baroni(2020)}]{baroni2020linguistic}
Marco Baroni. 2020.
\newblock Linguistic generalization and compositionality in modern artificial
  neural networks.
\newblock \emph{Philosophical Transactions of the Royal Society B},
  375(1791):20190307.

\bibitem[{Beaver(2001)}]{Beaver01}
David Beaver. 2001.
\newblock \emph{Presupposition and Assertion in Dynamic Semantics}.
\newblock CSLI Publications.

\bibitem[{Bowman et~al.(2015{\natexlab{a}})Bowman, Angeli, Potts, and
  Manning}]{snli:emnlp2015}
Samuel~R. Bowman, Gabor Angeli, Christopher Potts, and Christopher~D. Manning.
  2015{\natexlab{a}}.
\newblock A large annotated corpus for learning natural language inference.
\newblock In \emph{Proceedings of the 2015 Conference on Empirical Methods in
  Natural Language Processing (EMNLP)}, pages 632--642.

\bibitem[{Bowman et~al.(2015{\natexlab{b}})Bowman, Potts, and
  Manning}]{Bowman2015}
Samuel~R. Bowman, Christopher Potts, and Christopher~D. Manning.
  2015{\natexlab{b}}.
\newblock Recursive neural networks can learn logical semantics.
\newblock In \emph{Proceedings of the 3rd Workshop on Continuous Vector Space
  Models and their Compositionality}, pages 12--21.

\bibitem[{Clark et~al.(2020)Clark, Tafjord, and Richardson}]{Clark2020}
Peter Clark, Oyvind Tafjord, and Kyle Richardson. 2020.
\newblock Transformers as soft reasoners over language.
\newblock In \emph{Proceedings of the 29th International Joint Conference on
  Artificial Intelligence and the 17th Pacific Rim International Conference on
  Artificial Intelligence (IJCAI-PRICAI)}.

\bibitem[{Dagan et~al.(2013)Dagan, Roth, Sammons, and
  Zanzotto}]{series/synthesis/2013Dagan}
Ido Dagan, Dan Roth, Mark Sammons, and Fabio~Massimo Zanzotto. 2013.
\newblock \emph{Recognizing Textual Entailment: Models and Applications}.
\newblock Synthesis Lectures on Human Language Technologies. Morgan \& Claypool
  Publishers.

\bibitem[{van Dalen(2013)}]{vanDalen2013}
Dirk van Dalen. 2013.
\newblock \emph{Logic and Structure}, 5 edition.
\newblock Springer.

\bibitem[{Dasgupta et~al.(2018)Dasgupta, Guo, Stuhlm\"{u}ller, Gershman, and
  Goodman}]{dasgupta2018evaluating}
Ishita Dasgupta, Demi Guo, Andreas Stuhlm\"{u}ller, Samuel~J. Gershman, and
  Noah~D. Goodman. 2018.
\newblock {E}valuating compositionality in sentence embeddings.
\newblock In \emph{Proceedings of the 40th Annual Conference of the Cognitive
  Science Society}, pages 1596--1601.

\bibitem[{Devlin et~al.(2019)Devlin, Ming-Wei, Kenton, and
  Kristina}]{BERT2018new}
Jacob Devlin, Chang Ming-Wei, Lee Kenton, and Toutanova Kristina. 2019.
\newblock {BERT}: Pre-training of deep bidirectional transformers for language
  understanding.
\newblock In \emph{Proceedings of the 2019 Conference of the North American
  Chapter of the Association for Computational Linguistics: Human Language
  Technologies (NAACL-HLT)}, pages 4171--4186.

\bibitem[{Fodor and Pylyshyn(1988)}]{Fodor1988-FODCAC}
Jerry~A. Fodor and Zenon~W. Pylyshyn. 1988.
\newblock Connectionism and cognitive architecture: A critical analysis.
\newblock \emph{Cognition}, 28(1-2):3--71.

\bibitem[{Geiger et~al.(2019)Geiger, Cases, Karttunen, and Potts}]{Geiger2019}
Atticus Geiger, Ignacio Cases, Lauri Karttunen, and Christopher Potts. 2019.
\newblock Posing fair generalization tasks for natural language inference.
\newblock In \emph{Proceedings of the 2019 Conference on Empirical Methods in
  Natural Language Processing and the 9th International Joint Conference on
  Natural Language Processing (EMNLP-IJCNLP)}, pages 4484--4494.

\bibitem[{Geiger et~al.(2020)Geiger, Richardson, and
  Potts}]{geiger-etal-2020-neural}
Atticus Geiger, Kyle Richardson, and Christopher Potts. 2020.
\newblock Neural natural language inference models partially embed theories of
  lexical entailment and negation.
\newblock In \emph{Proceedings of the Third BlackboxNLP Workshop on Analyzing
  and Interpreting Neural Networks for NLP}, pages 163--173.

\bibitem[{Glockner et~al.(2018)Glockner, Shwartz, and
  Goldberg}]{glockner-shwartz-goldberg:2018:Short}
Max Glockner, Vered Shwartz, and Yoav Goldberg. 2018.
\newblock Breaking {NLI} systems with sentences that require simple lexical
  inferences.
\newblock In \emph{Proceedings of the 56th Annual Meeting of the Association
  for Computational Linguistics (ACL)}, pages 650--655.

\bibitem[{Goodwin et~al.(2020)Goodwin, Sinha, and
  O{'}Donnell}]{goodwin-etal-2020-probing}
Emily Goodwin, Koustuv Sinha, and Timothy~J. O{'}Donnell. 2020.
\newblock Probing linguistic systematicity.
\newblock In \emph{Proceedings of the 58th Annual Meeting of the Association
  for Computational Linguistics (ACL)}, pages 1958--1969.

\bibitem[{Gururangan et~al.(2018)Gururangan, Swayamdipta, Levy, Schwartz,
  Bowman, and Smith}]{gururangan-EtAl:2018:N18-2}
Suchin Gururangan, Swabha Swayamdipta, Omer Levy, Roy Schwartz, Samuel Bowman,
  and Noah~A. Smith. 2018.
\newblock Annotation artifacts in natural language inference data.
\newblock In \emph{Proceedings of the 2018 Conference of the North American
  Chapter of the Association for Computational Linguistics: Human Language
  Technologies (NAACL-HLT)}, pages 107--112.

\bibitem[{Heim and Kratzer(1998)}]{Heim-Kratzer98}
Irene Heim and Angelika Kratzer. 1998.
\newblock \emph{Semantics in Generative Grammar}.
\newblock Blackwell.

\bibitem[{Hochreiter and
  Schmidhuber(1997)}]{Hochreiter:1997:LSM:1246443.1246450}
Sepp Hochreiter and J\"{u}rgen Schmidhuber. 1997.
\newblock Long short-term memory.
\newblock \emph{Neural Comput.}, 9(8):1735--1780.

\bibitem[{Johnson-Laird and Byrne(1991)}]{johnson1991deduction}
Philip~N. Johnson-Laird and Ruth~M.J. Byrne. 1991.
\newblock \emph{Deduction}.
\newblock Erlbaum.

\bibitem[{Karttunen and Peters(1979)}]{Karttunen-Peters79}
Lauri Karttunen and Stanley Peters. 1979.
\newblock Conventional implicatures.
\newblock In Choon~Kyu Oh and David~A. Dineen, editors, \emph{Syntax and
  Semantics 11: Presupposition}, pages 1--56. Academic Press.

\bibitem[{Khemlani and Johnson-Laird(2012)}]{khemlani2012theories}
Sangeet Khemlani and Philip~N Johnson-Laird. 2012.
\newblock Theories of the syllogism: A meta-analysis.
\newblock \emph{Psychological bulletin}, 138(3):427--457.

\bibitem[{Kim and Linzen(2020)}]{kim-linzen-2020-cogs}
Najoung Kim and Tal Linzen. 2020.
\newblock {COGS}: A compositional generalization challenge based on semantic
  interpretation.
\newblock In \emph{Proceedings of the 2020 Conference on Empirical Methods in
  Natural Language Processing (EMNLP)}, pages 9087--9105.

\bibitem[{Kingma and Ba(2015)}]{Adam}
Diederik~P. Kingma and Jimmy Ba. 2015.
\newblock Adam: A method for stochastic optimization.
\newblock In \emph{Proceedings of the International Conference on Learning
  Representations (ICLR)}.

\bibitem[{Kuncoro et~al.(2018)Kuncoro, Dyer, Hale, Yogatama, Clark, and
  Blunsom}]{kuncoro-etal-2018-lstms}
Adhiguna Kuncoro, Chris Dyer, John Hale, Dani Yogatama, Stephen Clark, and Phil
  Blunsom. 2018.
\newblock {LSTM}s can learn syntax-sensitive dependencies well, but modeling
  structure makes them better.
\newblock In \emph{Proceedings of the 56th Annual Meeting of the Association
  for Computational Linguistics (ACL)}, pages 1426--1436.

\bibitem[{Lake and Baroni(2017)}]{Lake2017GeneralizationWS}
Brenden~M. Lake and Marco Baroni. 2017.
\newblock Generalization without systematicity: On the compositional skills of
  sequence-to-sequence recurrent networks.
\newblock In \emph{Proceedings of the International Conference on Machine
  Learning (ICML)}.

\bibitem[{Linzen et~al.(2016)Linzen, Dupoux, and
  Goldberg}]{linzen-etal-2016-assessing}
Tal Linzen, Emmanuel Dupoux, and Yoav Goldberg. 2016.
\newblock Assessing the ability of {LSTM}s to learn syntax-sensitive
  dependencies.
\newblock \emph{Transactions of the Association for Computational Linguistics
  (TACL)}, 4:521--535.

\bibitem[{Liu et~al.(2019)Liu, Schwartz, and Smith}]{liu-etal-2019-inoculation}
Nelson~F. Liu, Roy Schwartz, and Noah~A. Smith. 2019.
\newblock Inoculation by fine-tuning: A method for analyzing challenge
  datasets.
\newblock In \emph{Proceedings of the 2019 Conference of the North {A}merican
  Chapter of the Association for Computational Linguistics: Human Language
  Technologies (NAACL-HLT)}, pages 2171--2179.

\bibitem[{Marcus(2003)}]{962dc7dfb35547148019f194381d2cc6}
Gary Marcus. 2003.
\newblock \emph{The Algebraic Mind: Integrating Connectionism and Cognitive
  Science}.
\newblock MIT Press.

\bibitem[{Marelli et~al.(2014)Marelli, Menini, Baroni, Bentivogli, Bernardi,
  and Zamparelli}]{marelli-etal-2014-sick}
Marco Marelli, Stefano Menini, Marco Baroni, Luisa Bentivogli, Raffaella
  Bernardi, and Roberto Zamparelli. 2014.
\newblock A {SICK} cure for the evaluation of compositional distributional
  semantic models.
\newblock In \emph{Proceedings of the Ninth International Conference on
  Language Resources and Evaluation ({LREC})}, pages 216--223.

\bibitem[{de~Marneffe et~al.(2012)de~Marneffe, Manning, and
  Potts}]{de-marneffe-etal-2012-happen}
Marie-Catherine de~Marneffe, Christopher~D. Manning, and Christopher Potts.
  2012.
\newblock Did it happen? the pragmatic complexity of veridicality assessment.
\newblock \emph{Computational Linguistics}, 38(2):301--333.

\bibitem[{McCoy et~al.(2019)McCoy, Pavlick, and Linzen}]{mccoy2019}
R.~Thomas McCoy, Ellie Pavlick, and Tal Linzen. 2019.
\newblock Right for the wrong reasons: Diagnosing syntactic heuristics in
  natural language inference.
\newblock In \emph{Proceedings of the 57th Annual Meeting of the Association
  for Computational Linguistics (ACL)}, pages 3428--3448.

\bibitem[{Min et~al.(2020)Min, McCoy, Das, Pitler, and
  Linzen}]{min2020syntactic}
Junghyun Min, R.~Thomas McCoy, Dipanjan Das, Emily Pitler, and Tal Linzen.
  2020.
\newblock Syntactic data augmentation increases robustness to inference
  heuristics.
\newblock In \emph{Proceedings of the 58th Annual Meeting of the Association
  for Computational Linguistics (ACL)}, pages 2339--2352.

\bibitem[{Mineshima et~al.(2015)Mineshima, Mart\'{i}nez-G\'{o}mez, Miyao, and
  Bekki}]{D16-1242}
Koji Mineshima, Pascual Mart\'{i}nez-G\'{o}mez, Yusuke Miyao, and Daisuke
  Bekki. 2015.
\newblock Higher-order logical inference with compositional semantics.
\newblock In \emph{Proceedings of the 2015 Conference on Empirical Methods in
  Natural Language Processing (EMNLP)}, pages 2055--2061.

\bibitem[{Montague(1973)}]{Montague73}
Richard Montague. 1973.
\newblock The proper treatment of quantification in ordinary {E}nglish.
\newblock In Jaakko Hintikka, Julius M.~E. Moravcsik, and Patrick Suppes,
  editors, \emph{Approaches to Natural Language}, pages 189--224. Reidel,
  Dordrecht.
\newblock Reprinted in Richmond H. Thomason (ed.), \textit{{F}ormal
  {P}hilosophy: {S}elected {P}apers of {R}ichard {M}ontague}, 247--270, 1974,
  New Haven: Yale University Press.

\bibitem[{Nie et~al.(2019)Nie, Wang, and Bansal}]{DBLP:conf/aaai/NieWB19}
Yixin Nie, Yicheng Wang, and Mohit Bansal. 2019.
\newblock Analyzing compositionality-sensitivity of {NLI} models.
\newblock In \emph{Proceedings of the AAAI Conference on Artificial
  Intelligence}, pages 6867--6874.

\bibitem[{Ning et~al.(2017)Ning, Feng, and Roth}]{ning-etal-2017-structured}
Qiang Ning, Zhili Feng, and Dan Roth. 2017.
\newblock A structured learning approach to temporal relation extraction.
\newblock In \emph{Proceedings of the 2017 Conference on Empirical Methods in
  Natural Language Processing (EMNLP)}, pages 1027--1037.

\bibitem[{Pennington et~al.(2014)Pennington, Socher, and
  Manning}]{pennington-etal-2014-glove}
Jeffrey Pennington, Richard Socher, and Christopher Manning. 2014.
\newblock {G}lo{V}e: Global vectors for word representation.
\newblock In \emph{Proceedings of the 2014 Conference on Empirical Methods in
  Natural Language Processing ({EMNLP})}, pages 1532--1543.

\bibitem[{Poliak et~al.(2018)Poliak, Naradowsky, Haldar, Rudinger, and
  Van~Durme}]{poliak-EtAl:2018:S18-2}
Adam Poliak, Jason Naradowsky, Aparajita Haldar, Rachel Rudinger, and Benjamin
  Van~Durme. 2018.
\newblock Hypothesis only baselines in natural language inference.
\newblock In \emph{Proceedings of the Seventh Joint Conference on Lexical and
  Computational Semantics (*SEM)}, pages 180--191.

\bibitem[{Richardson et~al.(2020)Richardson, Hu, Moss, and
  Sabharwal}]{Richardson2019}
Kyle Richardson, Hai Hu, Lawrence~S. Moss, and Ashish Sabharwal. 2020.
\newblock Probing natural language inference models through semantic fragments.
\newblock In \emph{Proceedings of the AAAI Conference on Artificial
  Intelligence}.

\bibitem[{Ross and Pavlick(2019)}]{ross-pavlick-2019-well}
Alexis Ross and Ellie Pavlick. 2019.
\newblock How well do {NLI} models capture verb veridicality?
\newblock In \emph{Proceedings of the 2019 Conference on Empirical Methods in
  Natural Language Processing and the 9th International Joint Conference on
  Natural Language Processing (EMNLP-IJCNLP)}, pages 2230--2240.

\bibitem[{Rozen et~al.(2019)Rozen, Shwartz, Aharoni, and
  Dagan}]{rozen-etal-2019-diversify}
Ohad Rozen, Vered Shwartz, Roee Aharoni, and Ido Dagan. 2019.
\newblock Diversify your datasets: Analyzing generalization via controlled
  variance in adversarial datasets.
\newblock In \emph{Proceedings of the 23rd Conference on Computational Natural
  Language Learning (CoNLL)}, pages 196--205.

\bibitem[{Sinha et~al.(2019)Sinha, Sodhani, Dong, Pineau, and
  Hamilton}]{sinha-etal-2019-clutrr}
Koustuv Sinha, Shagun Sodhani, Jin Dong, Joelle Pineau, and William~L.
  Hamilton. 2019.
\newblock {CLUTRR}: A diagnostic benchmark for inductive reasoning from text.
\newblock In \emph{Proceedings of the 2019 Conference on Empirical Methods in
  Natural Language Processing and the 9th International Joint Conference on
  Natural Language Processing (EMNLP-IJCNLP)}, pages 4506--4515.

\bibitem[{Talmor and Berant(2019)}]{talmor-berant-2019-multiqa}
Alon Talmor and Jonathan Berant. 2019.
\newblock {M}ulti{QA}: An empirical investigation of generalization and
  transfer in reading comprehension.
\newblock In \emph{Proceedings of the 57th Annual Meeting of the Association
  for Computational Linguistics (ACL)}, pages 4911--4921.

\bibitem[{Troelstra and Schwichtenberg(2000)}]{Troelstra-Schwichtenberg00}
Anne~S. Troelstra and Helmut Schwichtenberg. 2000.
\newblock \emph{Basic Proof Theory}, 2 edition.
\newblock Cambridge University Press.

\bibitem[{Tsuchiya(2018)}]{DBLP:conf/lrec/Tsuchiya18}
Masatoshi Tsuchiya. 2018.
\newblock Performance impact caused by hidden bias of training data for
  recognizing textual entailment.
\newblock In \emph{Proceedings of the 11th International Conference on Language
  Resources and Evaluation (LREC)}.

\bibitem[{Wang et~al.(2019{\natexlab{a}})Wang, Pruksachatkun, Nangia, Singh,
  Michael, Hill, Levy, and Bowman}]{NIPS2019_8589}
Alex Wang, Yada Pruksachatkun, Nikita Nangia, Amanpreet Singh, Julian Michael,
  Felix Hill, Omer Levy, and Samuel Bowman. 2019{\natexlab{a}}.
\newblock {SuperGLUE}: A stickier benchmark for general-purpose language
  understanding systems.
\newblock In \emph{Proceedings of Advances in Neural Information Processing
  Systems 32 (NIPS)}, pages 3266--3280.

\bibitem[{Wang et~al.(2019{\natexlab{b}})Wang, Singh, Michael, Hill, Levy, and
  Bowman}]{wang2018glue}
Alex Wang, Amanpreet Singh, Julian Michael, Felix Hill, Omer Levy, and Samuel
  Bowman. 2019{\natexlab{b}}.
\newblock {GLUE}: A multi-task benchmark and analysis platform for natural
  language understanding.
\newblock In \emph{Proceedings of the International Conference on Learning
  Representations (ICLR)}.

\bibitem[{Weiss et~al.(2018)Weiss, Goldberg, and
  Yahav}]{weiss-etal-2018-practical}
Gail Weiss, Yoav Goldberg, and Eran Yahav. 2018.
\newblock On the practical computational power of finite precision {RNN}s for
  language recognition.
\newblock In \emph{Proceedings of the 56th Annual Meeting of the Association
  for Computational Linguistics (ACL)}, pages 740--745.

\bibitem[{White and Rawlins(2018)}]{white2018role}
Aaron~Steven White and Kyle Rawlins. 2018.
\newblock The role of veridicality and factivity in clause selection.
\newblock In \emph{Proceedings of the 48th Annual Meeting of the North East
  Linguistic Society, Amherst, MA, USA. GLSA Publications}.

\bibitem[{White et~al.(2018)White, Rudinger, Rawlins, and
  Van~Durme}]{white-etal-2018-lexicosyntactic}
Aaron~Steven White, Rachel Rudinger, Kyle Rawlins, and Benjamin Van~Durme.
  2018.
\newblock Lexicosyntactic inference in neural models.
\newblock In \emph{Proceedings of the 2018 Conference on Empirical Methods in
  Natural Language Processing (EMNLP)}, pages 4717--4724.

\bibitem[{Williams et~al.(2018)Williams, Nangia, and
  Bowman}]{DBLP:journals/corr/WilliamsNB17}
Adina Williams, Nikita Nangia, and Samuel Bowman. 2018.
\newblock A broad-coverage challenge corpus for sentence understanding through
  inference.
\newblock In \emph{Proceedings of the 2018 Conference of the North American
  Chapter of the Association for Computational Linguistics: Human Language
  Technologies (NAACL-HLT)}, pages 1112--1122.

\bibitem[{Yanaka et~al.(2020)Yanaka, Mineshima, Bekki, and
  Inui}]{yanaka-EtAl:2020:acl}
Hitomi Yanaka, Koji Mineshima, Daisuke Bekki, and Kentaro Inui. 2020.
\newblock Do neural models learn systematicity of monotonicity inference in
  natural language?
\newblock In \emph{Proceedings of the 58th Annual Meeting of the Association
  for Computational Linguistics (ACL)}, pages 6105--6117.

\bibitem[{Yanaka et~al.(2019{\natexlab{a}})Yanaka, Mineshima, Bekki, Inui,
  Sekine, Abzianidze, and Bos}]{yanaka-etal-2019-neural}
Hitomi Yanaka, Koji Mineshima, Daisuke Bekki, Kentaro Inui, Satoshi Sekine,
  Lasha Abzianidze, and Johan Bos. 2019{\natexlab{a}}.
\newblock Can neural networks understand monotonicity reasoning?
\newblock In \emph{Proceedings of the 2019 ACL Workshop BlackboxNLP: Analyzing
  and Interpreting Neural Networks for NLP}, pages 31--40.

\bibitem[{Yanaka et~al.(2019{\natexlab{b}})Yanaka, Mineshima, Bekki, Inui,
  Sekine, Abzianidze, and Bos}]{yanaka2019}
Hitomi Yanaka, Koji Mineshima, Daisuke Bekki, Kentaro Inui, Satoshi Sekine,
  Lasha Abzianidze, and Johan Bos. 2019{\natexlab{b}}.
\newblock {HELP}: A dataset for identifying shortcomings of neural models in
  monotonicity reasoning.
\newblock In \emph{Proceedings of the Eighth Joint Conference on Lexical and
  Computational Semantics (*{SEM})}, pages 250--255.

\bibitem[{Zhang et~al.(2017)Zhang, Rudinger, Duh, and
  Van~Durme}]{zhang-etal-2017-ordinal}
Sheng Zhang, Rachel Rudinger, Kevin Duh, and Benjamin Van~Durme. 2017.
\newblock Ordinal common-sense inference.
\newblock \emph{Transactions of the Association for Computational Linguistics
  (TACL)}, 5:379--395.

\end{thebibliography}
\clearpage
\appendix

\begin{table*}[!ht]
\centering
\scalebox{0.83}{
\begin{tabular}{ll} \hline
Syntax & Semantics \\ \hline 
$\catS \to \catNP \ \catVPpast$ & $\sem{\catS} = \sem{\catNP}(\sem{\catVPpast})$ \\
$\catS \to \catSNEG \ \catS$ & $\sem{\catS} = \sem{\catSNEG}(\sem{\catS})$ \\ 
$\catNP \to \catPN$ & $\sem{\catNP} = \sem{\catPN}$ \\
$\catNP \to \catPN \ \catConn \ \catPN$ & $\sem{\catNP} = \lambda F. \sem{\catConn}(\sem{\catPN}(F),\sem{\catPN}(F))$ \\
$\catNP \to \catPN \, , \, \catPN \, , \, \catConn \ \catPN$ & $\sem{\catNP} = \lambda F. \sem{\catConn}(\sem{\catPN}(F), \sem{\catConn}(\sem{\catPN}(F),\sem{\catPN}(F)))$ \\
$\catVPtense \to \catIVtense$ & $\sem{\catVPtense} = \sem{\catIVtense}$ \\
$\catVPtense \to \catTVtense \ \catNP$ & $\sem{\catVPtense} = \lambda x. \sem{\catNP}(\lambda y. \sem{\catTVtense}(x,y))$ \\
$\catVPpast \to \catVNEG \ \catVPbase$ & $\sem{\catVPpast} = \lambda x. \sem{\catVNEG}(\sem{\catVPbase}(x))$ \\
$\catPN \to \strcat{Ann} \mid \strcat{Bob} \mid \strcat{Chris} \mid \cdots$ & $\sem{\catPN} = \lambda F.F(\symb)$ \\
$\catIVbase \to \strcat{swim} \mid \strcat{drink} \mid \strcat{smoke} \mid \cdots$ & $\sem{\catIVbase} = \lambda x.\symb(x)$ \\
$\catIVpast \to \strcat{swam} \mid \strcat{drank} \mid \strcat{smoked} \mid \cdots$ & $\sem{\catIVpast} = \lambda x. \symb(x)$ \\
$\catTVbase \to \strcat{see} \mid \strcat{visit} \mid \strcat{touch} \mid \cdots$ & $\sem{\catTVbase} = \lambda y \lambda x. \symb(x,y)$ \\
$\catTVpast \to \strcat{saw} \mid \strcat{visited} \mid \strcat{touched} \mid \cdots$ & $\sem{\catTVpast} = \lambda y x. \symb(x,y)$ \\
$\catSNEG \to \strcat{it is not the case that}$ & $\sem{\catSNEG} = \lambda P.\neg P$ \\
$\catVNEG \to \strcat{did not}$ & $\sem{\catVNEG} = \lambda P.\neg P$ \\
$\catConn \to \strcat{and}$ & $\sem{\catConn} = \lambda P \lambda Q. P \wedge Q$ \\
$\catConn \to \strcat{or}$ & $\sem{\catConn} = \lambda P \lambda Q. P \vee Q$ \\ \hline
\end{tabular}
}%

\caption{Grammar for the Boolean logic fragment with semantic composition.
Feature \textit{tense} for VP is either ``base'' or ``past.''
In semantic composition, \symb\ is the place where
the symbol (lemma) for a lexical item appears.}
\label{tab:propfrag}
\end{table*}

\begin{table*}[h]
\centering
\scalebox{0.85}{
\begin{tabular}[t]{cc|c|cccc} \hline
\multicolumn{3}{c|}{\textbf{Data}} &\multicolumn{4}{c}{\textbf{Model}} \\
\fpp&\ph&\fph&\textbf{LSTM-T}&\textbf{LSTM-M\&T}&\textbf{BERT-T}&\textbf{BERT-M\&T}\\ \hline
\yes&\yes&\textbf{\yes}&$99.8\pm0.1$&$100.0\pm0.1$&$100.0\pm0.0$&$100.0\pm0.0$ \\
\rowcolor{LightCyan}
\yes&\unk&\textbf{\unk}&$99.3\pm0.0$&$99.7\pm0.1$&$100.0\pm0.0$&$100.0\pm0.0$ \\ 
\unk&\yes&\textbf{\unk}&$99.4\pm0.2$&$99.8\pm0.1$&$100.0\pm0.0$&$100.0\pm0.0$ \\
\unk&\unk&\textbf{\unk}&$99.6\pm0.1$&$100.0\pm0.1$&$100.0\pm0.0$&$100.0\pm0.0$ \\
\hline
\end{tabular}
}
\caption{Accuracies on the random train-test split of our fully synthetic transitivity dataset.
\textbf{-T} indicates a model trained with the train split of the transitivity inference set, and \textbf{-M\&T} indicates a model trained with MNLI mixed with the train split.}
\label{tab:randoms}

\scalebox{0.85}{
\begin{tabular}[t]{cc|c|cccc} \hline
\multicolumn{3}{c|}{\textbf{Data}} &\multicolumn{4}{c}{\textbf{Model}} \\
\fpp&\ph&\fph&\textbf{LSTM-T}&\textbf{LSTM-M\&T}&\textbf{BERT-T}&\textbf{BERT-M\&T}\\ \hline
\yes&\yes&\textbf{\yes}&$99.2\pm0.0$&$100.0\pm0.1$&$100.0\pm0.0$&$100.0\pm0.0$ \\
\rowcolor{LightCyan}
\yes&\unk&\textbf{\unk}&$98.4\pm0.2$&$99.1\pm0.1$&$100.0\pm0.0$&$100.0\pm0.0$ \\ 
\unk&\yes&\textbf{\unk}&$98.3\pm0.2$&$99.3\pm0.1$&$100.0\pm0.0$&$100.0\pm0.0$ \\
\unk&\unk&\textbf{\unk}&$99.6\pm0.1$&$100.0\pm0.1$&$100.0\pm0.0$&$100.0\pm0.0$ \\
\hline
\end{tabular}
}
\caption{Accuracies on the random train-test split of our naturalistic transitivity test set.
}
\label{tab:randomn}
\end{table*}

\section{Details about the Boolean logic fragment}

Table~\ref{tab:propfrag} shows the context-free grammar used to generate
sentences for Boolean logic reasoning with conjunction, disjunction, and negation.
Each rewriting rule is paired with the corresponding semantic composition rule in standard Montagovian semantics to generate the logical form of a sentence~\cite{Montague73,Heim-Kratzer98}.
We use ten items each for proper names (\catPN), intransitive verbs (\catIV), and transitive verbs (\catTV).
Each sentence is generated with a verb in the past tense.

For sentences with multiple NPs, we assume the surface-scope reading
where the subject NP takes scope over the object NP.
For instance,
the sentence \textit{Ann and Bob saw Chris or Daniel},
where the subject NP is conjunctive and the object NP is disjunctive, has the logical form
$(\LF{see}(\LF{ann}, \LF{chris}) \vee \LF{see}(\LF{ann}, \LF{daniel})) \wedge (\LF{see}(\LF{bob}, \LF{chris}) \vee \LF{see}(\LF{bob}, \LF{daniel}))$.

There are two types of negation, sentential negation (\catSNEG) and verbal negation (\catVNEG),
which are distinguished with respect to their scope interpretation. Thus, 
the sentence \textit{Ann and Bob did not swim} has the logical form
$\neg \LF{swim}(\LF{ann}) \wedge \neg \LF{swim}(\LF{bob})$, while
the sentence \textit{It is not the case that Ann and Bob did not swim} has the logical form
$\neg (\LF{swim}(\LF{ann}) \wedge \LF{swim}(\LF{bob}))$.

To generate a premise-hypothesis pair $(s_1, s_2)$ using this Boolean logic fragment,
we first generate a sentence $s_1$
and derive its logical form $F_1$ using the grammar in Table~\ref{tab:propfrag}.
We then randomly select one of the atomic formulas appearing in $F_1$, say $A$,
and takes its positive ($A$) or negative ($\neg A$) form, which is in turn converted to the hypothesis sentence $s_2$
using the same grammar.
The gold label (\textit{entailment} or \textit{non-entailment}) for the pair $(s_1, s_2)$ is determined by checking whether $F_1$ logically entails $A$ or $\neg A$
using a first-order-logic theorem prover\footnote{https://github.com/vprover/vampire}.

\section{Training details}
In all experiments, we trained models on eight NVIDIA DGX-1 Tesla V100 GPUs.
The runtime for training each model was about 1-8 hours, depending on the size of the training set.
The order of training instances was shuffled for each model.

\section{Supplementary results on the random train-test split}
To confirm that our transitivity inference dataset is not excessively difficult,
we conducted additional experiments using the random $9:1$ train$:$test split of transitivity inference (\fph)\ datasets.
We evaluate models under two settings: (i)~models trained with the train split of transitivity inference datasets and (ii)~models trained with the train split mixed with MNLI.
Table~\ref{tab:randoms} shows the results on the random train-test split of our full-synthetic transitivity dataset, and Table~\ref{tab:randomn} shows the results on the random train-test split of our naturalistic transitivity dataset.
These results showed that regardless of the existence of MNLI in the training set, models achieved perfect performance on our transitivity inference test set with the standard random train-test split setting.

\section{Human judgement details}
Using Amazon Mechanical Turk, we collected human judgements for 960 naturalistic veridical inference examples and 960 naturalistic transitivity inference examples.
We required raters to have completed at least 5,000 approved
tasks to maintain a 99\% approval rating.
Raters could indicate by a checkbox that one or both sentences did not make sense, but no rater clicked the checkbox.
We collected three annotations per pair and paid \$0.06 per labeled pair.

Since humans predict incorrect labels for some composite inference examples \fph\ where the verb \f\ is non-veridical,
we checked the accuracy of human judgement on a set of premise-hypothesis pairs \fpp\ and \fph\ involving each non-veridical verb, as shown in Table~\ref{tab:humanacc}.
Annotators tended to incorrectly make judgements for both \fpp\ and \fph.
Regarding accuracy for each non-veridical verb, annotators correctly drew inferences containing \textit{wish} and \textit{hope}, while they tended to draw inferences containing \textit{claim} and \textit{hear} incorrectly.

In comparison with the previous veridicality dataset MegaVeridicality2~\cite{white-etal-2018-lexicosyntactic}, the accuracy tended to be lower than that in
MegaVeridicality2\footnote{We calculated the percentage of the majority judgement for each verb for ten different annotations in MegaVeridicality2.}.
As (\ref{ex:megav2}) shows, while a simple complement is used for MegaVeridicality2, a natural complement like (\ref{ex:sickerra}) might induce veridicality bias~\cite{ross-pavlick-2019-well}, resulting in incorrect judgements on veridical inference.
Whether a verb is veridical or non-veridical, humans tend to judge the complement as true.

\begin{table}[h]
\scalebox{0.92}{
\begin{tabular}{lllll} \hline
         \textbf{Verb}&\fpp&\fph& \textbf{MegaV2} &  \\ \hline
argue    & 34 \diff{(-46)}                    & 66 \diff{(-14)}                   & 80                      &  \\
assume   & 70 \diff{(-15)}                   & 79 \diff{(-6)}                     & 85                      &  \\
believe  & 19 \diff{(-71)}                   & 59 \diff{(-31)}                   & 90                      &  \\
claim    & 15 \diff{(-65)}                   & 56 \diff{(-24)}                   & 80                      &  \\
doubt    & 91 \diff{(+9)}                   & 96 \diff{(+16)}                   & 80                      &  \\
estimate & 35 \diff{(-50)}                   & 64 \diff{(-21)}                    & 85                      &  \\
expect   & 53 \diff{(-27)}                   & 70 \diff{(-10)}                    & 80                      &  \\
feel     & 42 \diff{(-53)}                   & 67 \diff{(-28)}                   & 95                      &  \\
hear     & 14 \diff{(-41)}                   & 53 \diff{(-2)}                   & 55                      &  \\
hope     & 77 \diff{(-8)}                   & 92 \diff{(+7)}                   & 85                      &  \\
imply    & 18 \diff{(-47)}                   & 58 \diff{(-7)}                   & 65                      &  \\
predict  & 50 \diff{(-25)}                   & 73 \diff{(-2)}                   & 75                      &  \\
suspect  & 48 \diff{(-47)}                    & 79 \diff{(-16)}                   & 95                      &  \\
think    & 18 \diff{(-77)}                   & 57 \diff{(-38)}                   & 95                      &  \\
wish     & 77 \diff{(+7)}                    & 92 \diff{(+22)}                   & 70                      & \\ \hline
\end{tabular}
}
\caption{Accuracy (\%) of human judgements for each non-veridical verb. MegaV2 indicates the percentage of those annotators who judge each verb to be non-veridical
in MegaVeridicality2~\cite{white-etal-2018-lexicosyntactic}. \diff{A number in parentheses} is a difference from the accuracy in MegaVeridicality2.}
 \label{tab:humanacc}
\end{table}

\begin{exe}
\ex \label{ex:megav2}
\begin{xlist}
\exi{$f(s_1)$:} \label{ex:megav2a}
Someone \textbf{believed} that something happened.
\exi{$s_1$:} \label{ex:megav2b}
Something happened.
\end{xlist}
\ex \label{ex:sickerra}
\begin{xlist}
\exi{$f(s_1)$:} \label{ex:sickerraa}
Someone \textbf{believed} that a man is jumping off a low wall.
\exi{$s_1$:} \label{ex:sickerrab}
A man is jumping off a low wall.
\exi{$s_2$:} \label{ex:sickerrac}
A man is jumping a wall.
\end{xlist}
\end{exe}

\begin{figure*}[h!tb]
\begin{minipage}{0.5\hsize}
\subcaption{\ph, and a subset of \fph}
\scalebox{0.39}{\includegraphics[bb=0.000000 0.000000 576.001152 216.000432]{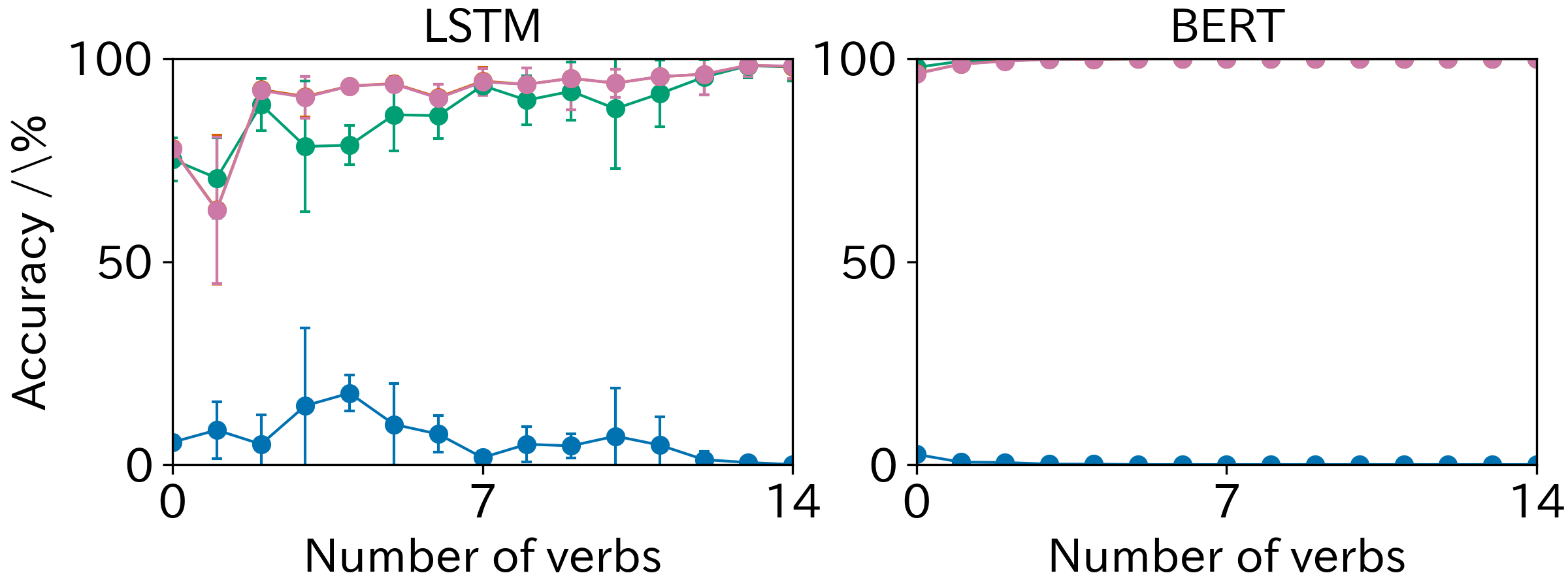}}
\end{minipage}
\begin{minipage}{0.5\hsize}
\subcaption{a subset of \fph}
\scalebox{0.39}{\includegraphics[bb=0.000000 0.000000 576.001152 216.000432]{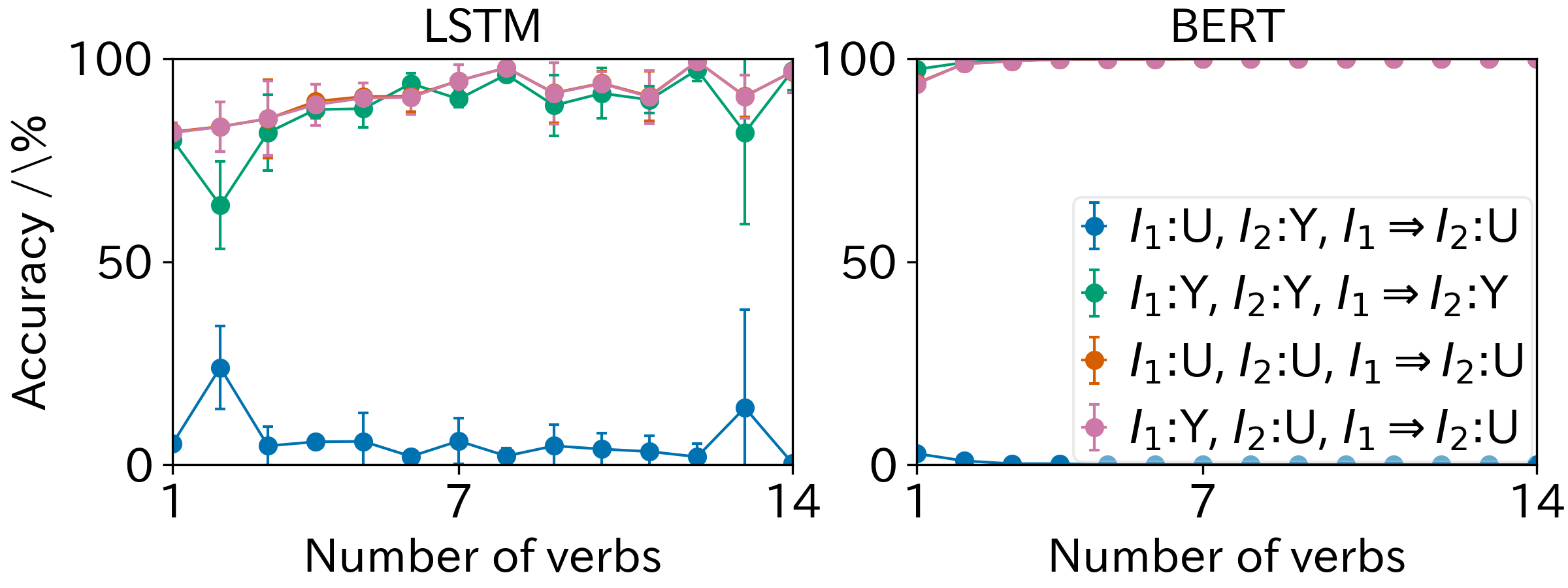}}
\end{minipage}
 \caption{Accuracies of models trained with (a) and (b). $I_1$ is the first basic pattern \fpp\ and $I_2$ is the second basic pattern \ph.
 Y indicates \textit{entailment} and U indicates \textit{non-entailment}. The horizontal axis shows the number of veridical verbs for the additional training set.}
 \label{fig:parta}
\end{figure*}

\section{Supplementary results with data augmentation}
In Section~\ref{ssec:dataaug}, we gradually added a subset of the composite inferences \fph\ involving a veridical verb (e.g., \textit{know}) to the training set and evaluated the performance of models on a naturalistic inference test set.
We also evaluated the performance of models under two conditions: (a)~models trained with the basic inference set \ph\ and a subset of the composite inference set \fph\ and (b) models trained with a subset of the composite inference set \fph.
Figure~\ref{fig:parta}(a) shows that the models significantly improved accuracy on composite inferences except for the test example \fph,
whose label differed from that of \ph.
Moreover, their performance was maintained even
without composite inference examples in the training set.
This indicates that models predict labels for the composite inference example only by judging whether it is similar to the basic inference example
in the training set.

Figure~\ref{fig:parta}(b) shows the result when models are trained only with a subset of the composite inference set \fph.
As non-veridical verbs are not included in the training set in this setting, the models predict labels for composite inferences involving non-veridical verbs by judging whether they are similar to composite inferences involving veridical verbs in the training set.
The models thus fail on composite inference examples \fph\ where \f\ is non-veridical and \ph\ is \yes.
The labels of such non-veridical inference examples are opposite to those of veridical inference examples in the training set.

\end{document}